\documentclass{article}

\usepackage{microtype}
\usepackage{graphicx}
\usepackage{subfigure}
\usepackage{booktabs} 

\usepackage{hyperref}

\usepackage[accepted]{icml2024}

\usepackage{amsmath}
\usepackage{amssymb}
\usepackage{mathtools}
\usepackage{amsthm}

\usepackage[capitalize,noabbrev]{cleveref}

\theoremstyle{plain}
\newtheorem{theorem}{Theorem}[section]

\newtheorem{lemma}[theorem]{Lemma}

\theoremstyle{definition}

\theoremstyle{remark}

\usepackage[textsize=tiny]{todonotes}

\DeclareMathOperator{\Ex}{\mathop{\mathbb{E}}}
\DeclareMathOperator{\Curv}{\mathrm{Curv}}

\DeclarePairedDelimiterX{\set}[1]{\{}{\}}{\setargs{#1}}
\NewDocumentCommand{\setargs}{>{\SplitArgument{1}{;}}m}
{\setargsaux#1}
\NewDocumentCommand{\setargsaux}{mm}
{\IfNoValueTF{#2}{#1} {#1\,\delimsize|\,\mathopen{}#2}}

\icmltitlerunning{Unveiling Privacy, Memorization, and Input Curvature Links}

\begin{document}

\twocolumn[
\icmltitle{Unveiling Privacy, Memorization, and Input Curvature Links}




\begin{icmlauthorlist}
\icmlauthor{Deepak Ravikumar}{ece}
\icmlauthor{Efstathia Soufleri}{ece} 
\icmlauthor{Abolfazl Hashemi}{ece}
\icmlauthor{Kaushik Roy}{ece}
\end{icmlauthorlist}

\icmlaffiliation{ece}{Department of ECE, Purdue University, West Lafayette, Indiana}

\icmlcorrespondingauthor{Deepak Ravikumar}{dravikum@purdue.edu}

\icmlkeywords{Machine Learning, ICML}

\vskip 0.3in
]



\printAffiliationsAndNotice{}  

\begin{abstract}
Deep Neural Nets (DNNs) have become a pervasive tool for solving many emerging problems.
However, they tend to overfit to and memorize the training set. 
Memorization 
is of keen interest since it is closely related to several concepts such as generalization, noisy learning, and privacy. To study memorization,
\citet{feldman2019does} proposed a formal score, however its computational requirements limit its practical use.
Recent research has shown empirical evidence linking input loss curvature (measured by the trace of the loss Hessian w.r.t inputs) and memorization. 
It was shown to be $\sim3$ orders of magnitude more efficient than calculating the memorization score. However, there is a lack of theoretical understanding linking memorization with input loss curvature.  
In this paper, we not only investigate this connection but also extend our analysis to establish theoretical links between differential privacy, memorization, and input loss curvature. First, we derive an upper bound on memorization characterized by both differential privacy and input loss curvature. Second, we present a novel insight showing that input loss curvature is upper-bounded by the differential privacy parameter. Our theoretical findings are further empirically validated using deep models on CIFAR and ImageNet datasets, showing a strong correlation between our theoretical predictions and results observed in practice.
\end{abstract}

\section{Introduction}
Machine learning and deep learning approaches have become state-of-the-art solutions in many learning tasks such as computer vision, natural language processing, etc. However, Deep Neural Nets (DNNs) are prone to over-fitting and memorization. An increasingly larger number of recent literature has focused on understanding memorization in DNNs \cite{zhang2017understanding, arpit2017closer, carlini2019distribution, feldman2019high, feldman2020neural, feldman2019does}. This is crucial given the implications to several connected areas such as generalization \cite{zhang2021understanding, brown2021memorization}, noisy learning \cite{liu2020early}, identifying mislabelled examples \cite{maini2022characterizing}, identifying rare and hard examples \cite{carlini2019distribution}, privacy \cite{feldman2019does}, risks from membership inference attacks \citep{shokri2017membership, carlini2022membership} and more. 

\begin{figure}
    \centering
    \includegraphics[width=0.85\columnwidth]{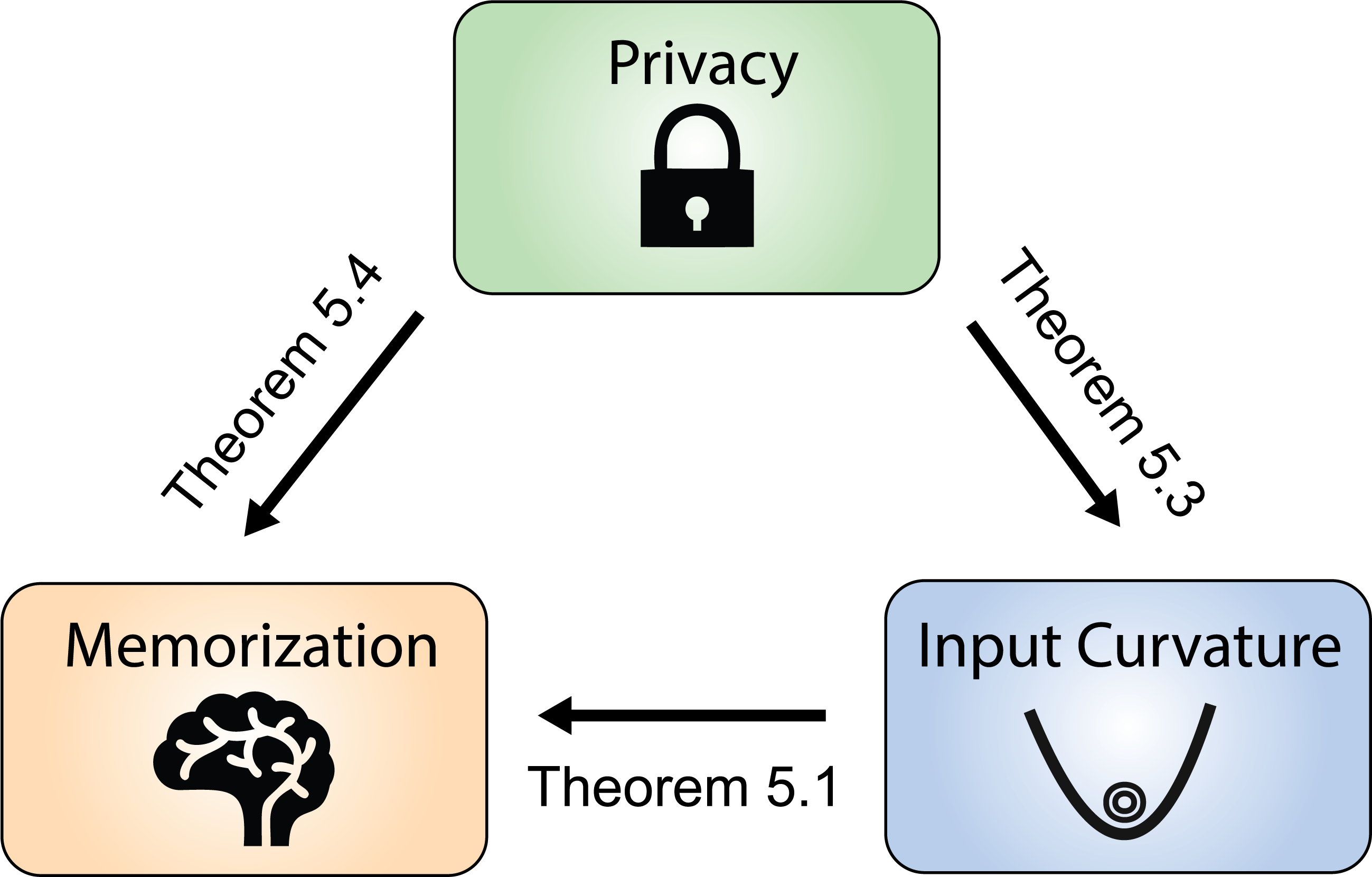}
    \caption{Our theoretical framework provides upper bounds in Theorems \ref{th:mem_curv},  \ref{th:pr_curv}, and \ref{th:pr_mem}. These are visualized as links between Differential Privacy, Memorization, and Input Loss Curvature.}
    \label{fig:intro-link}
    \vspace{-4mm}
\end{figure}

To study memorization several metrics have been suggested. \citet{carlini2019distribution} proposed a combination of five metrics to analyze memorization. Alternatively, \citet{jiang2020characterizing} proposed using a computationally efficient proxy to C-score, a metric closely related to the stability-based memorization \cite{feldman2019does}. The stability-based memorization score proposed by \citet{feldman2019does} measures the change in expected output probability when the sample under investigation is removed from the training dataset. Additionally, unlike other proposed metrics, \citet{feldman2019does} provides a strong theoretical framework for understanding memorization. This theory was then tested in practice in a subsequent paper \cite{feldman2020neural}. However, their method involved training thousands of models and is thus computationally infeasible in most real applications.

\begin{figure*}[hbt!]
\hfill
    \centering
    \subfigure[Low curvature examples from ImageNet.]{\includegraphics[width=0.8 \columnwidth]{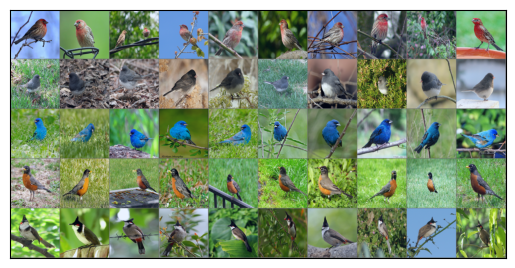}\label{fig:low_th_intro}}
    \vspace{-3mm}
    \hfill
    \subfigure[High curvature examples from ImageNet.]{\includegraphics[width=0.8 \columnwidth]{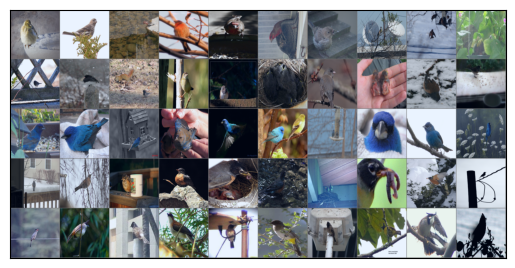}\label{fig:idkd_pre_post_dist}}
    \hfill
    \caption{Images from ImageNet ranked using input loss curvature. Input loss curvature was obtained using a single ResNet18 trained on ImageNet. Ten lowest curvature samples (left) and ten highest curvature samples (right) from the training set are visualized for 5 classes (each row is a class) from ImageNet. Low curvature samples are `prototypical' of their class, while high curvature samples are rare, difficult, and more likely memorized instances.}
    \label{fig:intro_figure}
    \vspace{-4mm}
\end{figure*}

In a recent paper, \citet{garg2023memorization} suggested a new proxy using input loss curvature to measure the stability-based memorization score proposed in \citet{feldman2019does}. To measure input loss curvature they suggested using the trace of the loss Hessian with respect to the input. Using this input loss curvature measurement, they provided empirical evidence on the link between memorization and input loss curvature. They obtained high cosine similarity between input loss curvature and memorization scores from \citet{feldman2020neural} while being $\sim3$ orders of magnitude less compute-intensive.
To illustrate the savings we reproduced \citet{garg2023memorization}'s results on ImageNet and visualized the ten lowest and highest curvature samples in Figure \ref{fig:intro_figure}. These examples were obtained using a \emph{single} ResNet18 model trained on ImageNet, compared to 1000's of models trained by \citet{feldman2020neural} to compute memorization scores. From Figure \ref{fig:intro_figure}, we see that low curvature samples are `prototypical' of their class. While high curvature samples are drawn from rare, hard, or outlier examples which are more likely to be memorized. 

Input loss curvature is thus, a promising proxy for stability-based memorization score. However, there is a lack of theoretical understanding of the link between memorization, and input loss curvature. In this paper, we develop a theoretical framework to understand this empirical observation and formally unveil the connections between memorization and input loss curvature. Further, we explore the relationship beyond memorization and input loss curvature, our theoretical contributions reveal the underlying link between differential privacy \citep{dwork2006calibrating}, memorization \cite{feldman2019does}, and input loss curvature \cite{garg2023memorization}. 
We present the links as three theorems. The first links memorization and input loss curvature, and the second theorem links input loss curvature and differential privacy. The third theorem links differential privacy and memorization. These links are visualized in Figure \ref{fig:intro-link}.
Each of the three theoretical links developed in this paper is corroborated by empirical evidence obtained on DNNs used for vision classification tasks.

In summary, the main contributions of this paper include:
\begin{itemize}
    \item We develop a theoretical framework for analyzing input loss curvature and memorization in a general learning setting and demonstrate its implications to DNNs.
    \item We show that memorization is upper bounded by (a) input loss curvature and (b) relevant privacy parameters. We also show that input loss curvature is also upper bounded by privacy, completing the theoretical links between memorization, privacy, and input loss curvature.
    \item We verify the theoretical results with extensive empirical experiments on vision classification tasks using DNNs on CIFAR100 and ImageNet datasets. 
    \item We obtain a tighter bound on private learnability. Namely, we establish that $\epsilon$-differential privacy implies $L(1-e^{-\epsilon})$ stability, thereby improving the previous theoretical bound.
\end{itemize}

\section{Preliminaries and Notation}
Consider a supervised learning problem where the goal is to learn a mapping from some input space $\mathcal{X} \subset \mathbb{R}^d$ to an output space $\mathcal{Y}\subset \mathbb{R}$. The learning is performed using a randomized algorithm $\mathcal{A}$ on a training set $S$. A randomized algorithm employs a degree of randomness as a part of its logic. 
The training set $S$ contains $m$ elements.
Each element $z_i = (x_i, y_i)$ is drawn from an unknown distribution $\mathcal{D}$, where $z_i \in \mathcal{Z}, x_i \in \mathcal{X}$, $y_i \in \mathcal{Y}$ and $\mathcal{Z} = \mathcal{X} \times \mathcal{Y}$. Thus we define the training set $S \in \mathcal{Z}^m$ as $S = \{z_1, \cdots, z_m\}$.
We assume $m \geq 2$. Another very relevant concept is adjacent datasets. Adjacent datasets are obtained when the $i^{th}$ element is removed. This is sometimes referred to as a leave-one-out set defined as
\begin{align*}
    S^{\setminus i} = \{z_1, \cdots, z_{i-1}, z_{i+1}, \cdots, z_m\}.
\end{align*}
Related to adjacent (a.k.a neighboring) datasets is the distance between datasets. The distance between any two datasets $S, S'$ denoted by $\lVert S - S' \rVert_1$ is a measure of how many samples differ between $S$ and $S'$. Note,  $\lVert S \rVert_1$ denotes the size of a dataset $S$.

A randomized learning algorithm $\mathcal{A}$ when applied on a dataset $S$ results in a hypothesis denoted by $h_{S}^\phi = \mathcal{A}(\phi, S),$
where $\phi \sim \Phi$ is the random variable associated with the randomness of the algorithm $\mathcal{A}$. A cost function
$c:\mathcal{Y}\times\mathcal{Y}\mapsto\mathbb{R}^+$ is used to measure the performance of the hypothesis. 
The cost of the hypothesis $h$ at a sample $z_i$ is also referred to as the loss $\ell$ at $z_i$ defined 
as 
\begin{align*}
  \ell(h, z_i) = c(h(x_i), y_i).  
\end{align*}
In most cases, we are interested in the loss of $h$ over the data distribution, which is referred to as population risk defined as
\begin{align*}
    R(h) = \Ex_{z\sim\mathcal{D}}[\ell(h, z)].
\end{align*}
Since the data distribution $\mathcal{D}$ is unknown in general, it is common to evaluate and study the empirical risk defined as
\begin{align*}
    R_{emp}(h, S) = \cfrac{1}{m} \sum_{i=1}^{m}\ell(h, z_i), \quad z_i \in S.
\end{align*}
In this paper, our characterization of curvature involves the gradient and the Hessian of the loss \textit{with respect to the input data}, which is denoted using the $\nabla$ and $\nabla^2$ operators respectively. $\lVert a \rVert$ denotes the $\ell_2$ norm of $a$.

\vspace{-2mm}
\section{Background}
\textbf{Differential Privacy} was introduced by \citet{dwork2006calibrating} and here we briefly recall the definition. 
A randomized algorithm $\mathcal{A}$ with domain $\mathcal{Z}^m$ is $\epsilon$-differentially private if for all $\mathcal{R} \subset \mathrm{Range} (\mathcal{A})$ and for all $S, S' \in \mathcal{Z}^m$ such that $||S-S'||_1 \leq 1$
\begin{align}
    \Pr_\phi[h^\phi_S \in \mathcal{R}] & \leq e^{\epsilon} \Pr_\phi[h^\phi_{S'} \in \mathcal{R}],
\label{eq:dp_def}
\end{align}
where the probability is taken over the randomness arising from the algorithm $\mathcal{A}, \phi \sim \Phi$. 

\textbf{Memorization} of the $i^{th}$ element $z_i = (x_i, y_i)$ of the dataset $S$ by an algorithm $\mathcal{A}$ was defined by \citet{feldman2019does} using the notion of stability as:
\begin{align}
    \mathrm{mem}(\mathcal{A}, S, i) &= \Pr_{\phi}[h_{S}^\phi(x_i) = y_i] - \Pr_{\phi}[h_{S^{\setminus i}}^\phi(x_i) = y_i],
\label{eq:mem}
\vspace{-4mm}
\end{align}
where the probability is taken over the randomness of algorithm $\mathcal{A}$.

\textbf{Error Stability}  of a possibly randomized algorithm $\mathcal{A}$ for some $\beta > 0$ is defined as \citet{kearns1997algorithmic} 
\begin{align}
     \forall i \in \{1, \cdots,  m\},~\left\lvert \Ex_{\phi, z}[\ell(h_S^\phi, z)] - \Ex_{\phi, z}[\ell(h^{\phi}_{S^{\setminus i}}, z)] \right\rvert~ \leq \beta,
\label{as:stability}
\end{align}
where $z \sim \mathcal{D}$ and $\phi \sim \Phi$.

\textbf{Generalization.} A randomized algorithm $\mathcal{A}$ is said to  generalize with confidence $\delta$ and  a rate $\gamma'(m)$ if
\begin{align}
\Pr[\left\lvert R_{emp}(h, S) - R(h) \right\rvert \leq \gamma'(m)] \geq \delta.
\label{as:gen}
\end{align}

\textbf{Uniform Model Bias.} The hypothesis $h$ resulting from the application of algorithm $\mathcal{A}$ to learn the true conditional $h^* = \Ex[y| x]$ from a dataset $S \sim \mathcal{D}^m$ has uniform bound on model bias given by $\Delta$ if
\begin{align}
    \forall S \sim \mathcal{D}^m,  \quad \left\lvert \Ex_\phi[R(h_S^\phi) - R(h^*)] \right \rvert \leq \Delta.
\label{as:model_bias}
\end{align}

\textbf{$\rho$-Lipschitz Hessian.} The Hessian of $\ell$ is Lipschitz continuous on $\mathcal{Z}$ if $\forall z_1, z_2 \in \mathcal{Z}$, and $\forall h \in \mathrm{Range}(\mathcal{A})$ if there exists some $\rho > 0$ such that
\begin{align}
\lVert \nabla^2_{z_1} \ell(h, z_1) - \nabla^2_{z_2} \ell(h, z_2) \rVert \leq \rho\lVert   z_1 - z_2 \rVert.
\label{as:hess_lip}
\end{align}

\textbf{Input Loss Curvature.} Using the notation of curvature from \citet{moosavi2019robustness, garg2023memorization}, input loss curvature is defined as the sum of the absolute eigenvalues of the Hessian $H$ of the loss with respect to input $z_i$, conveniently it can be written using the trace as
\begin{align}
    \Curv_{\phi}(z_i, S) = \mathrm{tr}(H) = \mathrm{tr}(\nabla^2_{z_i} \ell(h^{\phi}_S, z_i))
    \label{eq:def_curv}
\end{align}

\textbf{$\upsilon$-adjacency.} A dataset $S$ is said to contain $\upsilon$-adjacent (read as upsilon-adjacent) elements if it contains two elements $z_i, z_j$ such that $z_j = z_i + \alpha$ for some $\alpha \in B_p(\upsilon)$ (read as $\upsilon$-Ball). Note that this can be ensured through construction. Consider a dataset $S'$ which has no $z_j$ s.t $z_j = z_i + \alpha; z_j, z_i \in S'$. Then we can construct $S$  such that $S = \set{z ; z \in S'} \cup \set{z_i + \alpha}$ for some $z_i \in S', \alpha \in B_p(\upsilon)$, ensuring $\upsilon$-adjacency holds.

\section{Related Work}
\textbf{Input Loss Curvature} is a measure of the sensitivity of the model to a specific input. Loss curvature with respect to weight parameters has received significant attention \cite{keskar2017on, wu2020adversarial, Jiang2020Fantastic, foret2021sharpnessaware, kwon2021asam, andriushchenko2022towards}, recently regarding its role in characterizing the sharpness of a learning objective and its connection to generalization. However, input loss curvature has received less focus. Input loss curvature has been studied in the context of adversarial robustness \cite{fawzi2018empirical, moosavi2019robustness}, coresets \cite{garg2023samples} and recently as a proxy for memorization \cite{garg2023memorization}. 
\citet{moosavi2019robustness} showed that adversarial training decreases the curvature of the loss surface with respect to inputs. Further, they provided a theoretical link between robustness and curvature and proposed using curvature regularization. \citet{garg2023samples} identified samples with low curvature as being more data-efficient and developed a coreset identification and training algorithm based on input loss curvature. In an interesting application of input loss curvature, \citet{garg2023memorization} provided empirical evidence linking memorization and input loss curvature.

\textbf{Memorization} has garnered increasing research effort with several recent works aiming to add to the understanding of memorization and its implications \cite{zhang2017understanding, arpit2017closer, carlini2019distribution, feldman2019high, feldman2019does, feldman2020neural, maini2022characterizing, garg2023memorization, lukasik2023larger}. The motivation for studying memorization stems from a variety of goals ranging from deriving insights into generalization \cite{zhang2017understanding, toneva2018an, brown2021memorization, zhang2021understanding}, identifying mislabeled examples \cite{pleiss2020identifying, maini2022characterizing}, and identifying challenging or rare sub-populations \cite{carlini2019distribution}, to understanding privacy \cite{feldman2019does} and robustness risks from memorization \cite{shokri2017membership, carlini2022membership}. While several metrics have been proposed to study memorization \cite{carlini2019distribution, jiang2020characterizing}, the stability-based memorization score proposed by \citet{feldman2019does} provides a strong theoretical framework to understand memorization along with strong empirical evidence \cite{feldman2020neural}. However, since the score proposed by \citet{feldman2019does} is computationally expensive,   \citet{garg2023memorization} proposed using input loss curvature as a more compute-efficient proxy. In this paper, we develop the theoretical framework to understand the links between input loss curvature, memorization, and differential privacy.

\textbf{Influence Functions} were applied to deep learning by \citet{koh2017understanding} and are closely related to memorization. Influence functions aim to identify the impact of one training point on the model predictions. 
Influence functions try to answer the counterfactual: what would have happened
if a training point were absent, or if its values were changed slightly? 
While recent approaches \citep{schioppa2022scaling} have applied influence functions to large deep models, influence functions have been criticized \citep{basu2021influence, bae2022if, schioppa2023theoretical} since the underlying theory assumes strong convexity and positive definiteness of the Hessian, conditions that are not met in the context of DNNs. On the other hand, the theoretical framework we present in this paper does not make any assumptions about the convexity or the definiteness of the Hessian and is more suitable for studying deep learning.

\section{Linking Privacy, Memorization and Input Loss Curvature}

In this section, we discuss our theoretical contributions as three links. First, we present Theorem \ref{th:mem_curv} which links memorization and curvature. Second, we present Theorem \ref{th:pr_curv} which links privacy and curvature. Finally, we present Theorem \ref{th:pr_mem} which links memorization and privacy.

\vspace{-2mm}
\subsection{Memorization and Input Loss Curvature}
The association between memorization and input loss curvature may initially appear counterintuitive at first, but a closer examination reveals a fundamental connection. Both metrics intrinsically capture the sensitivity of a model to input perturbations. Here we provide a theoretical link between memorization and input curvature in the form of Theorem \ref{th:mem_curv}. Theorem \ref{th:mem_curv} is one of our core contributions. 

\begin{theorem}[Curvature Upper Bounds Memorization] \label{th:mem_curv} Let the assumptions of error stability \ref{as:stability}, generalization \ref{as:gen}, and uniform model bias \ref{as:model_bias} hold and assume the $\upsilon$-adjacency of the dataset and that the loss is bounded such that $0 \leq \ell \leq L$. Then with probability at least $1- \delta$  it holds
\begin{align}
   |\mathrm{mem}&(\mathcal{A}, S, i)| \leq  ~ \cfrac{1}{L} \Ex_{\phi}[\Curv_{\phi}(z_i, S^{\setminus i})] + c_1
   \label{eq:th_mem_curv}\\
   c_1= \cfrac{\rho}{6L}  & \Ex_\alpha[\lVert \alpha \rVert^3] + \cfrac{m\beta}{L} + \cfrac{(4m-1)\gamma}{L}  + \cfrac{2(m-1)\Delta}{L}
   \label{eq:mem_curv_offset}
\end{align}
\end{theorem}
\textit{Sketch of Proof.}
Using the result from \citet{nesterov2006cubic} we obtain an upper bound on the loss at $z_j$ involving the Hessian of the loss. By choosing $\alpha$ such that $\Ex [\alpha] = 0$ we get rid of the first-order terms. Then by taking expectation over the randomness of the algorithm and then performing algebraic manipulation we can show that the expected difference in loss at $z_i$ for two different models is upper bound by the result in Theorem \ref{th:mem_curv}. The final step is to make the connection that for bounded loss, the difference in loss at $z_i$ for two different models is a scaled version of memorization. The full proof is provided in Appendix \ref{sec:proof_mem_curv}.

\textbf{Interpreting the Theory.} 
Theorem \ref{th:mem_curv} (Equation \ref{eq:th_mem_curv}) indicates a \emph{ linear relationship} between memorization and input loss curvature. Observe that the upper bound is dependent on the input loss curvature of a sample $z_i$ and the offset factor $c_1$. However, the offset factor $c_1$ is data independent, i.e. $c_1$ has no dependence on $z_i$. The offset factor $c_1$ (Equation \ref{eq:mem_curv_offset}) consists of the following components. The third moment of the perturbation random variable $\alpha$, which is a measure of the skewness of the distribution. By choosing the distribution of $\alpha$ carefully, e.g. a centralized Gaussian, this can be made zero.  The second and third terms of Equation \ref{eq:mem_curv_offset} are properties of the training algorithm, i.e. the algorithm's stability $\beta$ and ability to generalize $\gamma$. The last term is dependent on model bias $\Delta$. Thus $c_1$ is roughly
\begin{align*}
    c_1 = \text{Stability} + \text{Generalization} + \text{Model Bias}
\end{align*}
To answer the question, `How tight is the upper bound?', we use empirical evaluation of curvature and memorization scores in Section \ref{sec:exp_curv_mem} and find that the \emph{linear relationship} from Equation \ref{eq:th_mem_curv} does hold true. We briefly and qualitatively discuss the validity of our assumptions in practical settings. Research \cite{hardt2016train} has shown that using stochastic gradient methods (such as stochastic gradient descent) to train models attains small generalization error. Further, it has been shown that stochastic gradient is uniformly
stable \cite{hardt2016train}. Thus the assumptions of stability (Equation \ref{as:stability}) and generalization (Equation \ref{as:gen}) are reasonable.  Model bias is a property of the model, and a uniform bound across different datasets seems reasonable. And finally, the $\upsilon$-adjacency can be ensured through construction.
In practice, this might not be needed because the size of the ball $B_p(\upsilon)$ is unconstrained. Thus, two samples from the same class that are `similar' may be sufficient to satisfy this requirement (note that this will result in a non-zero first term of Equation \ref{eq:mem_curv_offset}). With the size of modern datasets, this assumption is also reasonable.

\textbf{Remark.} Without assuming loss boundedness, we can state Theorem \ref{th:mem_curv} for cross-entropy replacing $L$ with $1$, if  
\begin{align*}
 \forall h \in \mathrm{Range} (\mathcal{A}), \forall k \quad    0 < \Pr[h(x_k) = y_k] < 1.
\end{align*}
Note the boundary condition that probability cannot be exactly 0 or 1. This is a reasonable assumption in a practical setting. The proof is provided in Appendix \ref{sec:proof_mem_curv_xe}. The main takeaway is that when the loss is bounded the expected difference in loss is the same as the memorization score, however when the loss is cross entropy the expected difference in loss upper bounds the memorization score.



\vspace{-2mm}
\subsection{Privacy and Input Loss Curvature}
In this section, we present the second link between input loss curvature and privacy. To make the connection between input curvature and privacy we leverage stability.  
To establish the curvature-privacy link we first present Lemma \ref{lm:stab_priv} which links the stability constant and privacy. In doing so, we further improve the bounds in \citet{wang2016learning}, from $L(e^{\epsilon} - 1)$ to $L(1 - e^{-\epsilon})$.
\begin{lemma}[Privacy $\implies$ Stability] \label{lm:stab_priv} Assume boundedness of the loss, i.e., $0 \leq \ell \leq L$. Then,  any $\epsilon$-differential
private algorithm satisfies $L(1 - e^{-\epsilon})$-stability. 
\end{lemma}

\textit{Sketch of Proof} We start with the difference in the expected loss of adjacent datasets. Next, we assume that models resulting from training on $S$ and $S^{\setminus i}$ for some $i$ have distributions $p$ and $p'$, respectively. We use the properties of the expectation operator to expand the resultant terms. Next, by upper bounding the expectation using loss boundedness and performing some algebraic manipulations we arrive at the result. The full proof is provided in Appendix \ref{sec:proof_stab_priv}.




Here we present our second main contribution in the form of Theorem \ref{th:pr_curv} linking privacy and input loss curvature.
\begin{theorem}[Privacy $\implies$ Low Input Loss Curvature] \label{th:pr_curv}  Let  $\mathcal{A}$ be a randomized algorithm which is $\epsilon$-differentially private and the assumptions of error stability \ref{as:stability}, generalization \ref{as:gen}, and uniform model bias \ref{as:model_bias} hold. Further, assume $0 \leq \ell \leq L$. Then for two adjacent datasets $S, S^{\setminus i} \sim \mathcal{D}$ with a probability at least $1-\delta$ we have 
\begin{align}
\Ex_{z,\phi}[\Curv_{\phi}(z, S)] \leq& ~L(m+1)(1- e^{-\epsilon}) + c_2 \label{eq:pr_curv}
\\
c_2 = (4m-1)\gamma &+ 2(m-1)\Delta + \cfrac{\rho}{6} \Ex[\lVert \alpha \rVert^3]
\end{align}
\end{theorem}
\textit{Sketch of Proof} 
Starting at Lemma \ref{lm:curv_intermidiate} we take the expectation over $z$, then we use the stability assumption. Rearranging the expressions and using Lemma \ref{lm:stab_priv} we arrive at the result. The full proof is provided in Appendix \ref{sec:proof_pr_curv}. 

\textbf{Interpreting Theorem \ref{th:pr_curv}.} Focusing on Equation \ref{eq:pr_curv}, we see that a stronger privacy guarantee ensures reduced average input loss curvature. To validate the tightness of the bound we use empirical evaluations of curvature and privacy in Section \ref{sec:exp_pr_curv}. Similar to Theorem \ref{th:mem_curv}, $c_2$ can be thought of as having two components, the generalization term $\gamma$ and the model bias term $\Delta$. By choosing $\alpha$ carefully (see previous discussion on Theorem \ref{th:mem_curv}) the last term can be ignored. Thus $c_2$ can be thought as $c_2 = \text{Generalization} + \text{Model Bias}$.
The validity of our assumptions in practical settings is reasonable as previously discussed for Theorem \ref{th:mem_curv}.

\vspace{-2mm}
\subsection{Privacy and Memorization}
The definition of stability-based memorization \cite{feldman2019does} is very much related to privacy. 
Notably, \citet{feldman2019does} explored this link, demonstrating that under specific conditions, algorithms that do not memorize cannot achieve optimal generalization performance. \citet{feldman2019does} showed that this memorization-generalization result stems from the long-tailed nature of data.
Our exploration in determining how memorization and privacy are related, is, however, different. In particular, we show that the memorization score is upper bounded by $1-e^{-\epsilon}$ for any $\epsilon$-DP algorithm. While this result is relatively straightforward, we state it for completeness as it is still a critical link in understanding memorization.
\begin{theorem}[Privacy $\implies$ Less Memorization] Let $\mathcal{A}$ be an $\epsilon$-differentially private algorithm and $z_i$ be the $i^{th}$ element of $S \in \mathcal{Z}^m$. Then, we have
\label{th:pr_mem}
\begin{align}
\forall i \in \{1, \cdots, m\}, \quad \mathrm{mem}(\mathcal{A}, S, i) &\leq 1 - e^{-\epsilon}.
\end{align}
\end{theorem}
\textit{Sketch of Proof} We start with the definition of $\epsilon$-differential privacy, with simple algebraic manipulation, and repetitively using the definition of $\epsilon$-differential privacy we arrive at the result. Note that this result can also be readily extended to ($\epsilon, \delta_p$)-differential privacy, i.e. Theorem \ref{th:pr_mem} holds for an ($\epsilon, \delta_p$)-differential private algorithm with a probability $1-\delta_p$. The full proof is provided in Appendix \ref{sec:proof_pr_mem}.

\vspace{-2mm}
\section{Experiments}
\subsection{Experimental Setup}
\label{sec:exp_setup}
\textbf{Datasets.} To evaluate our theory we consider the classification task using standard vision datasets as the pre-computed stability-based memorization scores from \citet{feldman2020neural} are available for CIFAR100 \cite{krizhevsky2009learning} and ImageNet \cite{ILSVRC15} datasets. 

\textbf{Architectures.} For some experiments we train ResNet18 \cite{he2016deep} models from scratch, while for others we use pre-trained Small Inception \cite{szegedy2015going} and ResNet50 models released by \citet{feldman2020neural}. Details regarding the model used are specified at the beginning of each experiment section. 

\textbf{Training.}
For experiments that use private models, we use the Opacus library \cite{opacus} to train ResNet18 models for 20 epochs till the privacy budget is reached. We use DP-SGD \citep{abadi2016deep} with the maximum gradient norm set to $1.0$ and privacy parameter $\delta=1\times 10^{-5}$.
The initial learning rate was set to $0.001$. The learning rate is decreased by $10$ at epochs $12$ and $16$. When training on CIFAR10 and CIFAR100 datasets the batch size is set to $128$. For both CIFAR10 and CIFAR100 datasets, we used the following sequence of data augmentations for training: resize ($32 \times 32$), random crop, and random horizontal flip, this is followed by normalization.

\textbf{Testing.} 
During testing no augmentations were used, i.e. we used resize followed by normalization. When using pre-trained models from \citet{feldman2020neural} we validated the accuracy of the models before performing experiments. To improve reproducibility, we have provided the code in the supplementary material.

\begin{figure*}[ht]
    \centering
    \subfigure{
        \includegraphics[scale=1]{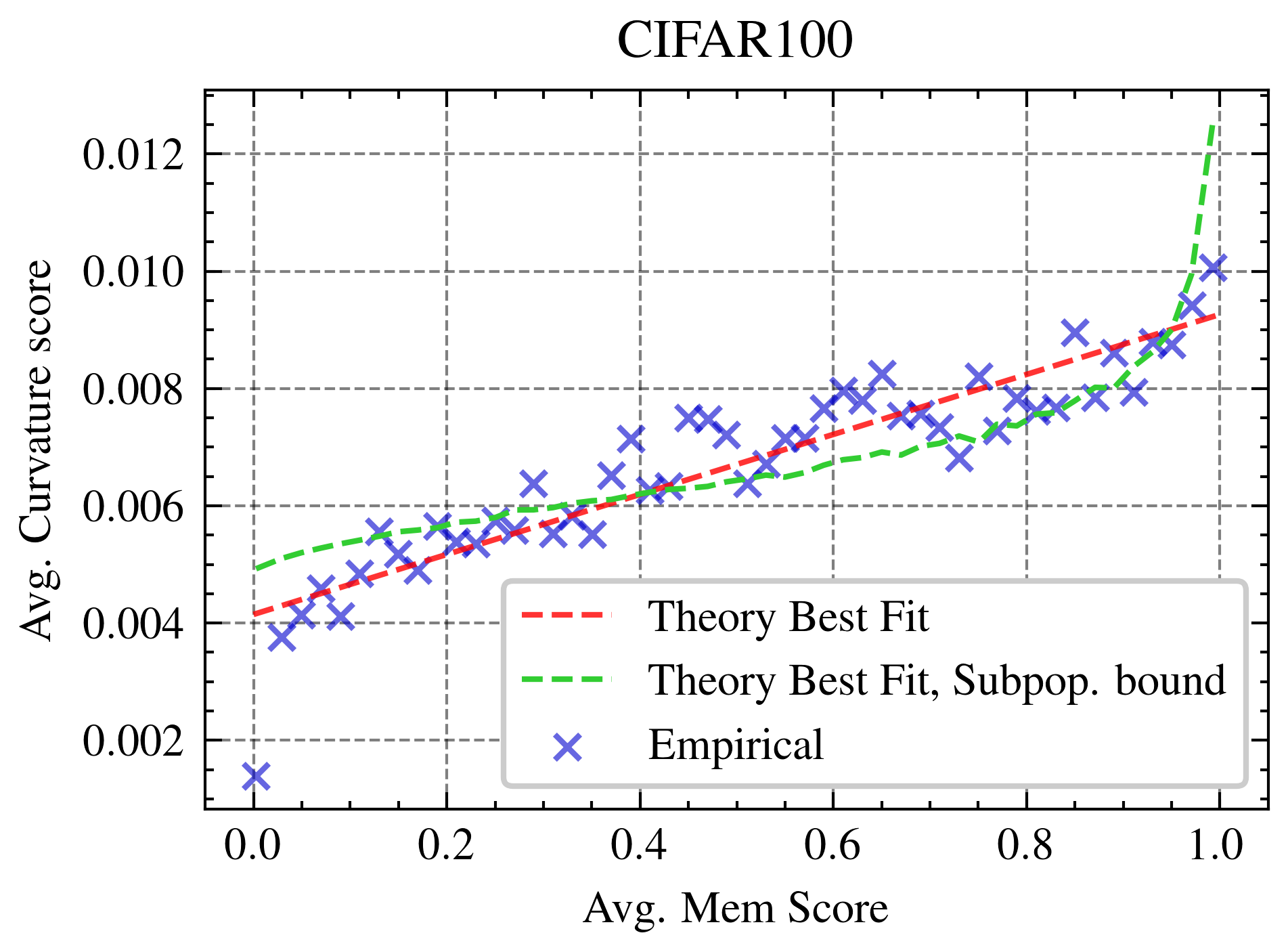}
        \label{fig:cifar100_fz_v_curv_1000}
    }
    \hfill
    \subfigure{
        \includegraphics[scale=1]{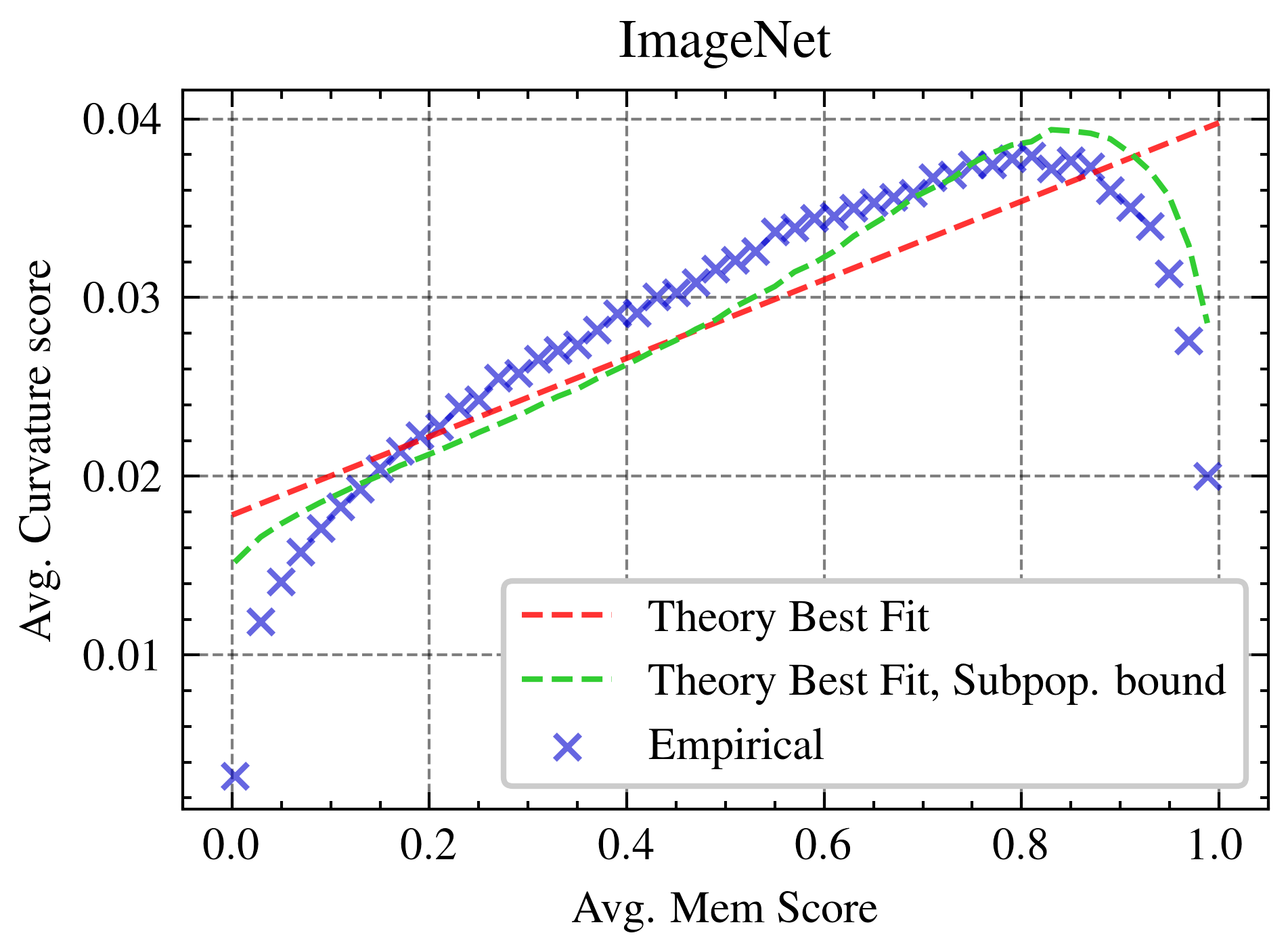}
        \label{fig:imgnet_fz_v_curv_100}
    }
    \vspace{-4mm}
    \caption{Plot of memorization score vs. input loss curvature at the end of training for CIFAR100 (average over 1000 Small Inception models) and ImageNet (average over 100 ResNet50) datasets.}
    \vspace{-4mm}
    \label{fig:multi_model}
\end{figure*}

\subsection{Estimating Input Loss Curvature}
To corroborate the theoretical findings presented in the prior section, an efficient methodology for computing input loss curvature is needed as computing the full Hessian is computationally intensive. Since we are interested in the trace of the Hessian, it can be efficiently computed using Hutchinson's trace estimator \cite{hutchinson1989stochastic, garg2023memorization} from which we have
\begin{align}
    \mathrm{tr}(H) = \mathbb{E}_v \left[  v^T H v   \right],
\end{align}
where $v \in \mathbb{R}^{d}$ belongs to a Rademacher distribution. Using the finite step approximation similar to \citet{moosavi2019robustness,garg2023memorization} and the symmetric nature of the Hessian we have
\begin{align}
    \mathrm{tr}(H^2) &= \frac{1}{n} \sum_{i=0}^{n} \rVert Hv_i \rVert_{2}^{2} \nonumber\\
    Hv &\propto \frac{\partial  \left(L(x+hv) - L(x)\right)}{\partial x} \nonumber\end{align}
\begin{align}\quad \mathrm{tr}(H^2) &\propto \frac{1}{n} \sum_{i=0}^{n} \left\lVert \frac{\partial  \left(L(x+hv) - L(x)\right)}{\partial x} \right\rVert^2_2 \nonumber \\
\Curv(x) &\propto \frac{1}{n} \sum_{i=0}^{n} \left\lVert \frac{\partial  \left(L(x+hv) - L(x)\right)}{\partial x} \right\rVert^2_2,
    \label{eq:curv}
\end{align}
where $n$ is the number of Rademacher vectors to average. For all our experiments we used $h=1\times10^{-3}$ and $n=10$. We found the results to be robust to changes in $h$; we varied it from $1\times10^{-1}$ to $1\times10^{-3}$. We also varied $n$ from $5,10,20$ and found the results to be robust to changes in $n$.


\vspace{-2mm}
\subsection{Input Curvature and Memorization}
\label{sec:exp_curv_mem}
In this section, we present the empirical results on CIFAR100 and ImageNet datasets for the first link between memorization and input loss curvature (Theorem \ref{th:mem_curv}). 

\textbf{Experiment.}
Here we aim to understand how memorization changes with curvature. The experiment aims to plot the memorization score vs curvature score to validate our theoretical results. 
We calculate curvature scores by averaging over many seeds at the end of training. This empirical measurement is proportional to the expected curvature score, i.e. $\Ex_{\phi}[\Curv_{\phi}(z_i, S^{\setminus i})]$ in Theorem \ref{th:mem_curv}.


For this experiment, we used 1000 models trained on CIFAR100 and 100 models trained on ImageNet obtained from \citet{feldman2020neural}'s 0.7 subset ratio repository. We calculated the curvature score for each sample in the training set using Equation \ref{eq:curv}. We then compiled a dataset comprising each sample's memorization score and curvature score. Precomputed memorization scores were obtained from \citet{feldman2020neural}'s repository.  We averaged these scores across all models (1000 for CIFAR100 and 100 for ImageNet) to form an averaged dataset, which was divided into 50 bins based on memorization score. For example, bin 0 includes samples with memorization scores from 0 to 0.02 and the corresponding curvature scores, bin 1 includes samples in the memorization score range of 0.02 to 0.04, and so on. The average memorization score and maximum curvature score (since curvature is an upper bound) for each bin were used to create a scatter plot as shown in Figures \ref{fig:cifar100_fz_v_curv_1000} and \ref{fig:imgnet_fz_v_curv_100}. For CIFAR100, the Small Inception model was used, and for ImageNet, the ResNet50 model was used, both sourced from  \citet{feldman2020neural}.


\textbf{Results.} We provide the results for CIFAR100 and ImageNet datasets in Figure \ref{fig:cifar100_fz_v_curv_1000} and \ref{fig:imgnet_fz_v_curv_100} respectively. The figures also visualize the best-fit  (shown in red) based on Theorem \ref{th:mem_curv}. From the results, we see a clear linear relation. The results from the experiment show that the curvature scores have a strong linear trend with respect to memorization, in line with Theorem \ref{th:mem_curv}. 

\textbf{Accounting for Variables in Practice.}
Notably, the linearity of the relation between memorization and curvature diminishes at the extreme ends of the data range. This phenomenon is particularly pronounced in ImageNet results, as shown in Figure \ref{fig:imgnet_fz_v_curv_100}. This is because the loss bound $L$ (refer to Equation \ref{eq:th_mem_curv}) is not constant and the bound changes for each sub-population. Here, we can treat each memorization bin as a sub-population. Hence, when using cross entropy loss we found a better fit, if the loss boundedness is assumed, and the loss bound at convergence is empirically modeled. Accounting for the change in loss bound with sub-population size we see a much improved match. This is observed when comparing the best-fit results in green (assuming sub-population loss bound) vs red (no sub-population loss bound) in Figure \ref{fig:multi_model}.

To obtain an improved fit seen in Figure \ref{fig:multi_model} we assumed the loss bound reduces in the square root of the sub-population size \citep{bousquet2002stability}. 
Since the theoretical curvature score from Equation \ref{eq:def_curv} is proportional to the computed curvature score (Equation \ref{eq:curv}), we can rewrite Equation \ref{eq:th_mem_curv} from Theorm \ref{th:mem_curv} using two parameters $p_1, c_1$ as
\begin{align}
|\mathrm{mem}&(\mathcal{A}, S, i)| \leq  ~ \cfrac{p_1}{L} \cdot \Ex_{\phi}[\Curv_{\phi}(z_i, S^{\setminus i})] + c_1 \nonumber
\end{align}
Using $L \propto m_{sub}^{-0.5}$, where $m_{sub}$ is the number of samples in each sub-population we can model the relation as
\begin{align}
    |\mathrm{mem}(\mathcal{A}, S, i)| \leq& ~ p_1 \cdot \sqrt{m_{sub}} \cdot \Ex_{\phi}[\Curv_{\phi}(z_i, S^{\setminus i})]   + c_1 \nonumber\\ \quad s.t. \quad & p_1, c_1 > 0.
\end{align}
Fitting parameters $p_1, c_1$ to the data results in the green plot in Figures \ref{fig:cifar100_fz_v_curv_1000} and \ref{fig:imgnet_fz_v_curv_100}, where we see much improved match between empirical results and our theory. Thus, these results strongly agree with and validate Theorem \ref{th:mem_curv}.

\subsection{Privacy and Input Loss Curvature}
\label{sec:exp_pr_curv}
In this section, we present the empirical results on CIFAR10 and CIFAR100 datasets to verify the link between privacy and input loss curvature (Theorem \ref{th:pr_curv}).

\textbf{Experiment.}
To study the relation between privacy and curvature, we train private ResNet18 models on CIFAR10 and CIFAR100 using DP-SGD \cite{abadi2016deep} and calculate the curvature scores. We aim to plot privacy budget vs curvature score and validate Theorem \ref{th:pr_curv}. Specifically, we train models with privacy budgets $\epsilon$ ranging from 5 to 100, in increments of 5. We train 10 seeds for every privacy budget, and the curvature score is averaged over the 10 seeds and all the dataset samples.

\begin{figure}[t!]
    \centering
    \includegraphics[scale=1]{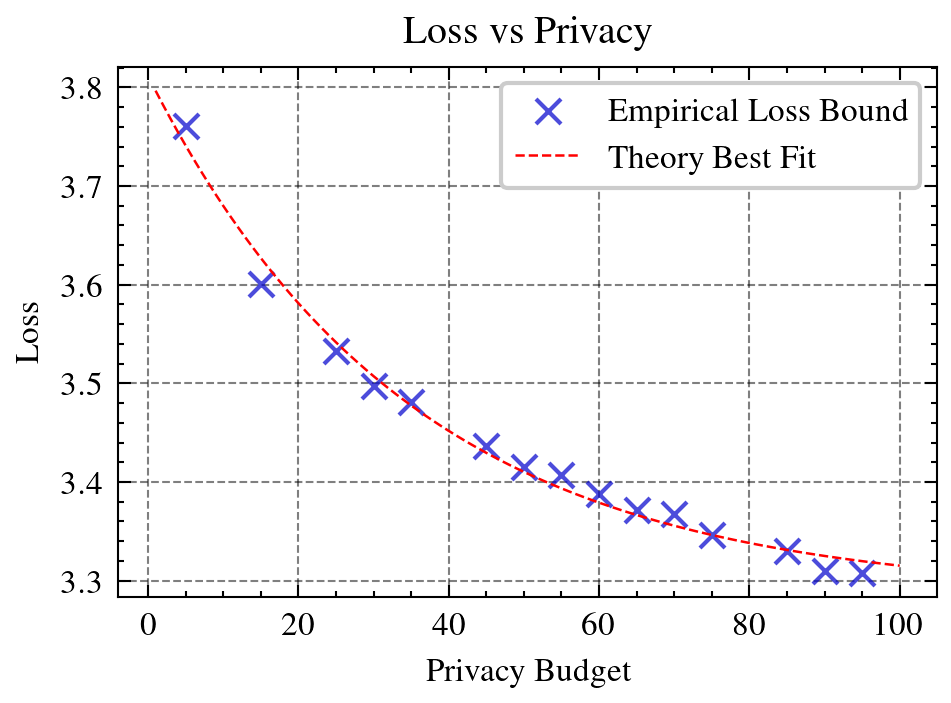}
    \vspace{-4mm}
    \caption{Plot of differential privacy vs loss bound for CIFAR100 trained with cross-entropy and the best fit curve (dashed).}
    \label{fig:pr_loss}
\end{figure}

\begin{figure}[t!]
    \centering
    \includegraphics[scale=1]{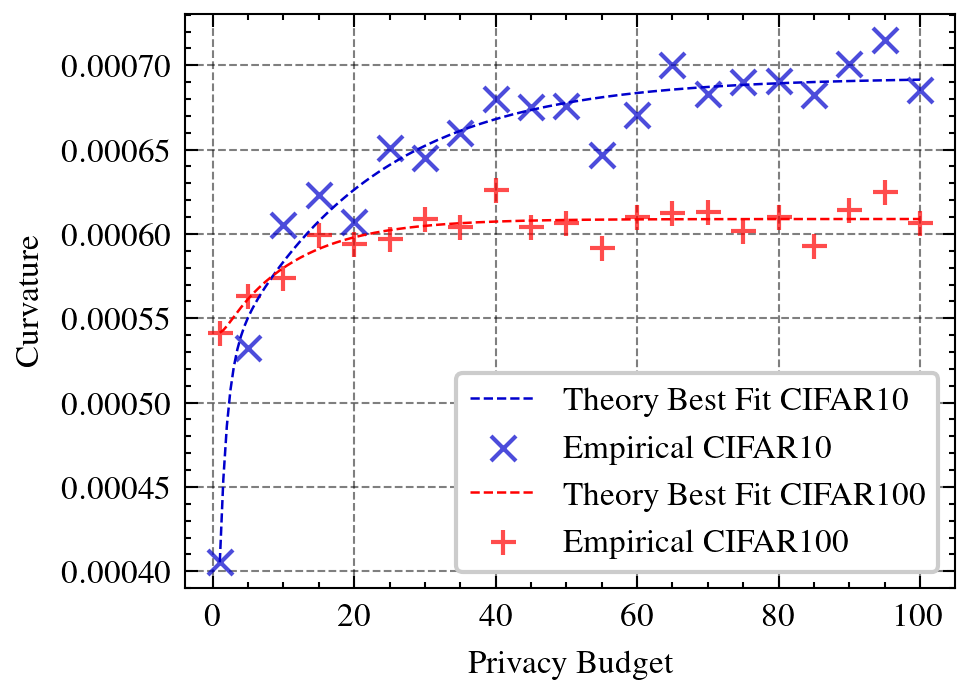}
    \vspace{-6mm}
    \caption{Plot of privacy vs loss curvature for CIFAR10 and CIFAR100.  The best-fit curve (dashed) is predicted by
    Theorem \ref{th:pr_curv}.} 
    \label{fig:pr_curv}
    \vspace{-5mm}
\end{figure}

\textbf{Accounting for Variables in Practice.}
For our experiments, we use cross entropy trained private models, where the loss is unbounded. However, Theorem \ref{th:pr_curv} assumes bounded loss. Thus, we obtain an empirical bound on the loss at convergence for each privacy budget. We model the loss bound as a function of privacy using $L(\epsilon) = a + b e^{-c  \epsilon}$.  The fit of this model is shown in Figure \ref{fig:pr_loss}. Using this loss bound model, Theorem \ref{th:pr_curv} can be re-written as
\begin{align}
    \Ex_{z,\phi}&[\Curv_{\phi}(z, S)] \leq L(\epsilon) \cdot (m + 1) \cdot (1 - e^{-\epsilon}) + c_2 \nonumber\\
    \quad &\leq  (a + b e^{-c  \epsilon}) \cdot (m + 1) \cdot (1 - e^{-\epsilon}) + c_2,    \label{eq:model_curv_xe}
\end{align}
where $c_2$ is treated as a constant when trying to fit the data to Equation \ref{eq:model_curv_xe}. The empirical data and the best fit model using Equation \ref{eq:model_curv_xe} are shown in Figure \ref{fig:pr_curv}. 

\textbf{Results.}
The result of plotting the average convergence loss and privacy budget is shown in Figure \ref{fig:pr_loss} along with the best-fit model (in dashed blue line), demonstrating a strong match. Next, Figure \ref{fig:pr_curv} shows the result of studying the link between input loss curvature and privacy budget. The scatter plot shows curvature vs privacy. We visualize the empirical data and the best fit (dashed line) using the model from Equation \ref{eq:model_curv_xe}. Again, we see a very strong match. All these results strongly correlate with theory and validate Theorem \ref{th:pr_curv}.

\subsection{Memorization and Privacy}
In this section, we present the results for the link between memorization and privacy (Theorem \ref{th:pr_mem}). 

\textbf{Experiment.}
The goal of the study is to estimate the memorization score of samples when the models have differential privacy guarantees. 
Since Theorem \ref{th:pr_mem}  provides an upper bound, we are interested in how privacy affects most memorized examples. This enables us to reduce the compute requirement, and we consider the top 500 most memorized samples from CIFAR100 as reported in \citet{feldman2020neural} and study how these scores change as privacy guarantees are varied.
For this experiment we first split the CIFAR100 training set into two, set $a$ contains all examples that are not the top 500 most memorized examples, and set $b$ contains the top 500 most memorized examples as reported by \citet{feldman2020neural}. From $b$ we randomly sample half the dataset called $b^{0.5}$. We concatenate $a$ and $b^{0.5}$ to get our training set. This is used to train a ResNet18 model using DP-SGD \citep[Differentially Private SGD]{abadi2016deep}. We repeat the process of random sub-sampling of $b$ and training for 40 seeds. By keeping track of what samples of $b$ were present in each training run we can estimate the memorization score of the top 500 most memorized examples. This process is repeated 6 times for privacy budgets $\epsilon=1,10,20,30,40,50$ with $\delta = 1  \times 10^{-5}$ to train a total of 240 private models (previously described in Section \ref{sec:exp_setup}). 

\begin{figure}[t!]
    \centering
    \includegraphics[scale=1]{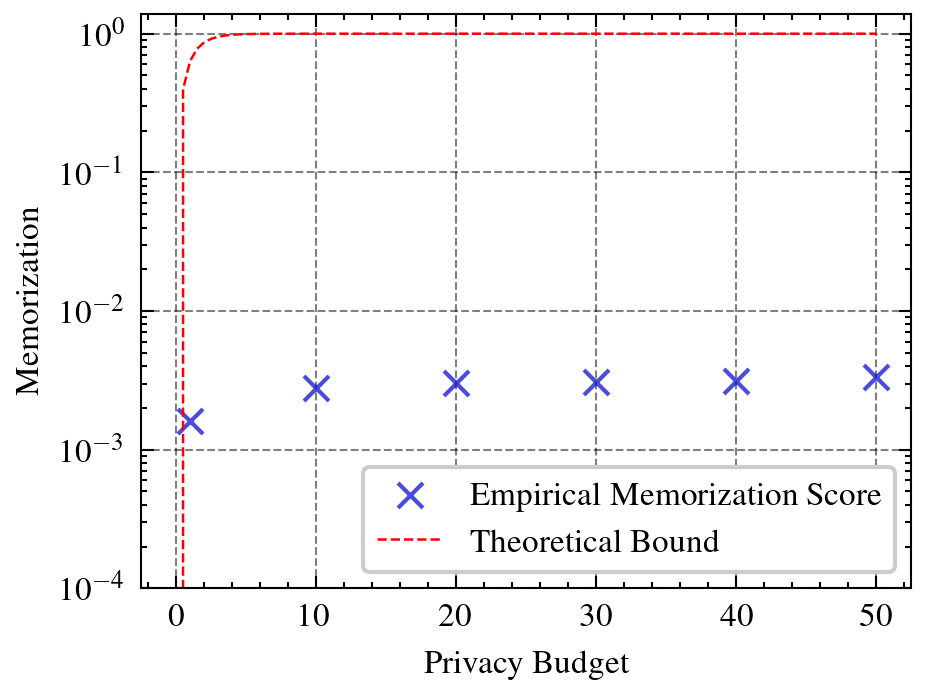}
    \vspace{-4mm}
    \caption{Plot of differential privacy vs memorization for CIFAR100 and the upper bound from the Theorem \ref{th:pr_mem}.}
    \label{fig:pr_mem}
    \vspace{-3mm}
\end{figure}

\textbf{Results.}
The average memorization scores for the top 500 most memorized examples across various privacy budgets ($\epsilon$) are presented in Figure \ref{fig:pr_mem}. As a reference, we also plot the upper bound from Theorem \ref{th:pr_mem} in the same plot. Note that Figure \ref{fig:pr_mem} is a semi-log plot. The results align with Theorem \ref{th:pr_mem}, showing an increase in memorization score as the privacy budget increases (i.e. privacy budget $\epsilon \uparrow$). Further, the memorization scores are significantly lower than the bound from Theorem \ref{th:pr_mem} supporting \citet{nasr2021adversary}'s observation that DP-SGD may be overly conservative.

\section{Conclusion}
This paper explores the theoretical link between memorization, curvature, and privacy. Understanding this link is critical since input curvature offers $\sim3$ orders of magnitude compute efficiency when calculating memorization scores. The theoretical analysis relies on three assumptions, stability, generalization, and Lipshitzness, and thus can be applied in non-convex settings such as DNNs. Our main result shows that memorization is upper-bounded by curvature and privacy. Further, we presented two theorems that complete the links between memorization, privacy, and input loss curvature.  To empirically test the theory we use standard DNNs for image classification using CIFAR100 and ImageNet datasets. Our results show a very strong match between our theoretical findings and empirical results. Results in this paper provide evidence for the link between memorization, input loss curvature, and privacy strengthening the understanding of DNNs and their properties.




\newpage
\section*{Impact Statement}
The research presented in this paper fills significant gaps in our understanding of DNNs. We focused on the relationship between memorization, input loss curvature, and privacy.
This finding is key for various applications, as it provides a clearer framework for leveraging the significant ($\sim3$ orders of magnitude) efficiencies in computing memorization scores when using input loss curvature. 
This work, therefore, not only develops our theoretical understanding of DNNs but also offers practical insights for developing more effective machine learning and deep learning models and algorithms.
\bibliography{ref}
\bibliographystyle{icml2024}

\newpage
\appendix
\onecolumn
\section{Proofs}

\subsection{Proof of Theorem \ref{th:pr_mem}} 
\label{sec:proof_pr_mem}
Consider $S, S^{\setminus i}$ from construction we have $||S-S^{\setminus i} || = 1$. Next let $\mathcal{R} \subset \mathrm{Range}(\mathcal{A})$ such that $\mathcal{R} = \{h ~\lvert~ h(x_i) = y_i\}$. Since $\mathcal{A}$ is $\epsilon$-differentially private then it follows from the definition of differential privacy in Equation \ref{eq:dp_def} that
\begin{align}
\Pr_\phi[h^\phi_S \in \mathcal{R}] & \leq e^{\epsilon} \Pr_\phi[h^\phi_{S\setminus i} \in \mathcal{R}]
\label{eq:dp_priv_mem}
\end{align}
Since $\mathcal{R} = \{h ~\lvert~ h(x_i) = y_i\}$ we have 
\begin{align}
    \Pr_{\phi}[h_S^\phi \in \mathcal{R}] = \Pr_{\phi}[h_S^\phi(x_i) = y_i]
\label{eq:r_hyp}
\end{align}
Using Equations \ref{eq:dp_priv_mem} and \ref{eq:r_hyp} we have
\begin{align}
\Pr_{\phi}[h^\phi_S(x_i) = y_i] & \leq e^{\epsilon}\Pr_{\phi}[h^\phi_{S^{\setminus i}}(x_i) = y_i]
\label{eq:long_notation} \\
\Pr_{\phi}[h^\phi_S(x_i) = y_i] & \leq e^{\epsilon} \Pr_{\phi}[h^\phi_{S^{\setminus i}}(x_i) = y_i] \pm \Pr_{\phi}[h^\phi_{S^{\setminus i}}(x_i) = y_i] \nonumber \\
\Pr_{\phi}[h^\phi_S(x_i) = y_i] - \Pr_{\phi}[h^\phi_{S^{\setminus i}}(x_i) = y_i]& \leq e^{\epsilon}\Pr_{\phi}[h^\phi_{S^{\setminus i}}(x_i) = y_i] - \Pr_{\phi}[h^\phi_{S^{\setminus i}}(x_i) = y_i] \nonumber \\
\Pr_{\phi}[h^\phi_S(x_i) = y_i] - \Pr_{\phi}[h^\phi_{S^{\setminus i}}(x_i) = y_i] & \leq (e^{\epsilon} - 1) \Pr_{\phi}[h^\phi_{S^{\setminus i}}(x_i) = y_i] \nonumber \\
\mathrm{mem}(\mathcal{A}, S, i) & \leq (e^{\epsilon} - 1) \Pr_{\phi}[h^\phi_{S^{\setminus i}}(x_i) = y_i] \nonumber 
\end{align}
Using Equation \ref{eq:long_notation}, we have the lower bound on $\Pr_{\phi}[h^\phi_{S^{\setminus i}}(x_i) = y_i]$ as 
\begin{align*}
    \Pr_{\phi}[h^\phi_{S^{\setminus i}}(x_i) = y_i] \geq e^{-\epsilon}\Pr_{\phi}[h^\phi_S(x_i) = y_i]
\end{align*}
Thus we have
\begin{align*}
    \mathrm{mem}(\mathcal{A}, S, i) &\leq (e^\epsilon - 1) e^{-\epsilon}\Pr_{\phi}[h^\phi_S(x_i) = y_i] \\
    \mathrm{mem}(\mathcal{A}, S, i) &\leq (1 - e^{-\epsilon})\Pr_{\phi}[h^\phi_S(x_i) = y_i]
\end{align*}
Since $\sup \Pr_{\phi}[h^\phi_S(x_i) = y_i] = 1$ we have the result
\begin{align*}
    \mathrm{mem}(\mathcal{A}, S, i) &\leq 1 - e^{-\epsilon} \quad \blacksquare
\end{align*}

\subsection{Proof of Lemma \ref{lm:loss_diff}}
\label{sec:proof_loss_diff}

\begin{lemma}
\label{lm:adj_gen_bound}
If the generalization assumption \ref{as:gen} holds then we know here exists $\gamma$ such that with probability $1-\delta$
\begin{align}
\Ex_\phi[\lvert R_{emp}(h_{S^{\setminus i}}^\phi, S) - R(h_{S^{\setminus i}}^\phi) \rvert] ~\leq \gamma \label{eq:adj_gen_1}\\
\Ex_\phi[\lvert R_{emp}(h_S^\phi, S) - R(h_S^\phi) \rvert] ~\leq \gamma \label{eq:adj_gen_2} \\
 \Ex_\phi[\lvert R_{emp}(h_{S}^\phi, S^{\setminus i}) - R(h_{S}^\phi) \rvert] ~\leq \gamma \label{eq:adj_gen_3} 
 \end{align}
\end{lemma}
\textbf{Proof of Lemma \ref{lm:adj_gen_bound}} From \cite{feldman2019high} we know that with a confidence $\delta$ we have 
\begin{align*}
    \Pr_{S\sim \mathcal{D}^m}\left[\left\lvert R_{emp}(h, S) - R(h) \right\rvert \geq c\left( \beta'\ln(m) \ln(m/\delta) + \cfrac{\sqrt{\ln(1/\delta)}}{\sqrt{m}} \right) \right] \leq \delta
\end{align*}
Where $\beta'$ is the uniform stability bound.
Thus with a confidence of at least $1 - \delta$ we can say:
\begin{align*}
    \left\lvert R_{emp}(h, S) - R(h) \right\rvert < c\left( \beta'\ln(m) \ln(m/\delta) + \cfrac{\sqrt{\ln(1/\delta)}}{\sqrt{m}} \right)
\end{align*}

Thus if we set
\begin{align*}
    \gamma'(m) = c\left( \beta'\ln(m) \ln(m/\delta) + \cfrac{\sqrt{\ln(1/\delta)}}{\sqrt{m}} \right)
\end{align*}
we have
\begin{align}
    \left\lvert R_{emp}(h, S) - R(h) \right\rvert < \gamma'(m)
    \label{eq:gen_bound_feldman}
\end{align}
Thus as a direct consequence of Equation \ref{eq:gen_bound_feldman} we can say
\begin{align}
 \forall S,S^{\setminus i} \sim \mathcal{D}^m,~ \Ex_\phi[\lvert R_{emp}(h_{S^{\setminus i}}^\phi, S^{\setminus i}) - R(h_{S^{\setminus i}}^\phi) \rvert] ~&\leq \gamma'(m-1) \\
  \forall S \sim \mathcal{D}^m,~\Ex_\phi[\lvert R_{emp}(h_S^\phi, S) - R(h_S^\phi) \rvert] ~&\leq \gamma'(m)
\end{align}
\begin{align*}
    \Ex_\phi[\lvert R_{emp}(h_{S^{\setminus i}}^\phi, S) - R(h_{S^{\setminus i}}^\phi) \rvert] &= \Ex_\phi\left[\left\lvert\cfrac{1}{m}\ell\left(h_{S^{\setminus i}}^\phi, z_i\right)\right\rvert\right] +\Ex_\phi\left[\left\lvert \cfrac{m-1}{m}R_{emp}(h_{S^{\setminus i}}^\phi, S^{\setminus i}) - R(h_{S^{\setminus i}}^\phi) \right\rvert\right]\\
    &= \cfrac{1}{m}\Ex_\phi\left[\left\lvert\ell\left(h_{S^{\setminus i}}^\phi, z_i\right)\right\rvert\right]  + \Ex_\phi\left[\left\lvert  \cfrac{m-1}{m}R_{emp}(h_{S^{\setminus i}}^\phi, S^{\setminus i}) -  \cfrac{m-1}{m}R(h_{S^{\setminus i}}^\phi) - \cfrac{1}{m}R(h_{S^{\setminus i}}^\phi)\right\rvert\right]\\
    &\leq \cfrac{L}{m} + \Ex_\phi\left[\left\lvert  \cfrac{m-1}{m}R_{emp}(h_{S^{\setminus i}}^\phi, S^{\setminus i}) -  \cfrac{m-1}{m}R(h_{S^{\setminus i}}^\phi) - \cfrac{1}{m}R(h_{S^{\setminus i}}^\phi)\right\rvert\right]\\
    &\leq \cfrac{L}{m} + \Ex_\phi\left[\left\lvert  \cfrac{m-1}{m}R_{emp}(h_{S^{\setminus i}}^\phi, S^{\setminus i}) -  \cfrac{m-1}{m}R(h_{S^{\setminus i}}^\phi) - \cfrac{1}{m}R(h_{S^{\setminus i}}^\phi) \pm \cfrac{1}{m} R(h^*)\right\rvert\right]\\
    &\leq \cfrac{L}{m} + \Ex_\phi\left[\left\lvert  \cfrac{m-1}{m}R_{emp}(h_{S^{\setminus i}}^\phi, S^{\setminus i}) -  \cfrac{m-1}{m}R(h_{S^{\setminus i}}^\phi) - \cfrac{1}{m} R(h^*)\right\rvert\right] + \Delta\\
    &\leq \cfrac{L}{m} + \Ex_\phi\left[\left\lvert  \cfrac{m-1}{m}R_{emp}(h_{S^{\setminus i}}^\phi, S^{\setminus i}) -  \cfrac{m-1}{m}R(h_{S^{\setminus i}}^\phi) - \cfrac{1}{m} R(h^*)\right\rvert\right] + \Delta\\
    &\leq \cfrac{L}{m} + \left\lvert\cfrac{m-1}{m}\gamma'(m-1)\right\rvert + \left\lvert\cfrac{1}{m} R(h^*)\right\rvert + \Delta\\
    &\leq \cfrac{L}{m} + \left\lvert\cfrac{m-1}{m}\gamma'(m-1)\right\rvert + \cfrac{L}{m} + \Delta\\
    &\leq \cfrac{2L}{m} + \cfrac{m-1}{m}\gamma'(m-1) + \Delta
\end{align*}
Now consider
\begin{align*}
    R_{emp}(h_S^\phi, S^{\setminus i}) - R(h_S^\phi)&=   R_{emp}(h_S^\phi, S^{\setminus i}) - R(h_S^\phi)\\
    &= \cfrac{1}{m-1} \sum_{j=1, j\neq i}^{m} \ell(h_S^\phi, z_j) - R(h_S^\phi)\\
    &= \cfrac{1}{m-1} \sum_{j=1}^{m} \ell(h_S^\phi, z_j) - \cfrac{1}{m-1} \ell(h_S^\phi, z_i) - R(h_S^\phi)\\
    &\leq \cfrac{m}{m-1} R_{emp}(h_S^\phi, S) - R(h_S^\phi)\\
    &\leq \gamma'(m) + \cfrac{1}{m-1} R_{emp}(h_S^\phi, S)\\
    &\leq \gamma'(m) + \cfrac{1}{m-1} L\\
    \lvert R_{emp}(h_S^\phi, S^{\setminus i}) - R(h_S^\phi) \rvert &\leq \gamma'(m) + \cfrac{L}{m-1}
\end{align*}
Thus if we set $\gamma(m) = \max \left\{\cfrac{2L}{m} + \cfrac{m-1}{m}\gamma'(m-1) + \Delta,\gamma'(m) + \cfrac{L}{m-1} \right\}$ we get $\forall S,S^{\setminus i} \sim \mathcal{D}^m$
\begin{align}
\Ex_\phi[\lvert R_{emp}(h_{S^{\setminus i}}^\phi, S) - R(h_{S^{\setminus i}}^\phi) \rvert] ~\leq \gamma \\
\Ex_\phi[\lvert R_{emp}(h_S^\phi, S) - R(h_S^\phi) \rvert] ~\leq \gamma
 \\
\Ex_\phi[\lvert R_{emp}(h_{S}^\phi, S^{\setminus i}) - R(h_{S}^\phi) \rvert] ~\leq \gamma
\end{align}

Using error stability from assumption (see Equation \ref{as:stability}) introduced by \citet{kearns1997algorithmic} without loss of generality, we can write
\begin{align*}
\Ex_{\phi, z \sim \mathcal{D}}[\ell(h_S^\phi, z)] - \Ex_{\phi, z\sim \mathcal{D}}[\ell(h_{S^{\setminus i}}^{\phi}, z)] &\leq \beta\\
\Ex_{\phi}[R(h_S^\phi) - R(h_{S^{\setminus i}}^{\phi})] &\leq \beta
\end{align*}

\begin{lemma}
\label{lm:loss_diff}
If assumptions of error stability \ref{as:stability}, generalization \ref{as:gen}, and uniform model bias \ref{as:model_bias} hold, then for all $ i, j$ and two adjacent datasets $S, S^{\setminus i} \sim \mathcal{D}$ with a probability at least $ 1-\delta$ it holds that 
\begin{align}
\left\lvert \Ex_{\phi}[\ell(h_S^\phi, z_i) - \ell(h_{S^{\setminus i}}^{\phi}, z_j)] \right\rvert \leq&~m\beta + (4m-1)\gamma \nonumber
\\& + 2(m-1)\Delta.
\end{align}
\end{lemma}
Lemma \ref{lm:loss_diff} provides an upper bound on the expected loss difference between two adjacent datasets evaluated at any data point in the training set. 

\textbf{Proof of Lemma \ref{lm:loss_diff}}
Using Lemma \ref{lm:adj_gen_bound} we know here exists $\gamma$ such that with probability $1-\delta$
\begin{align*}
\Ex_\phi[\lvert R_{emp}(h_{S^{\setminus i}}^\phi, S) - R(h_{S^{\setminus i}}^\phi) \rvert] ~\leq \gamma \\
\Ex_\phi[\lvert R_{emp}(h_S^\phi, S) - R(h_S^\phi) \rvert] ~\leq \gamma 
 \\
\Ex_\phi[\lvert R_{emp}(h_{S}^\phi, S^{\setminus i}) - R(h_{S}^\phi) \rvert] ~\leq \gamma
\end{align*}
Using Equations \ref{eq:adj_gen_1} and \ref{eq:adj_gen_2} we can upper bound the expected difference in empirical risk of adjacent datasets as
\begin{align*}
\Ex_{\phi}[R_{emp}(h_S^\phi, S) - R_{emp}(h_{S^{\setminus i}}^{\phi}, S)] &\leq \beta + 2\gamma\\
\Ex_{\phi}\left[\cfrac{1}{m}\ell(h_S^\phi, z_i) + \cfrac{m-1}{m}R_{emp}(h_S^\phi, S^{\setminus i}) - \cfrac{1}{m}\ell(h_{S^{\setminus i}}^{\phi}, z_j) - \cfrac{m-1}{m}R_{emp}(h_{S^{\setminus i}}^{\phi}, S^{\setminus i}) \right] &\leq \beta + 2\gamma
\end{align*}
\begin{align*}
\cfrac{1}{m}\Ex_{\phi}[\ell(h_S^\phi, z_i)] - \cfrac{1}{m}\Ex_{\phi}[\ell(h_{S^{\setminus i}}^{\phi}, z_j)] &\leq \beta + 2\gamma + \cfrac{m-1}{m}\Ex_{\phi}[R_{emp}(h_{S^{\setminus i}}^{\phi}, S^{\setminus i})  - R_{emp}(h_S^\phi, S^{\setminus i})]\\
\Ex_{\phi}[\ell(h_S^\phi, z_i)] - \Ex_{\phi}[\ell(h_{S^{\setminus i}}^{\phi}, z_j)] &\leq m\beta + 2m\gamma + (m-1)\Ex_{\phi}[R_{emp}(h_{S^{\setminus i}}^{\phi}, S^{\setminus i})  - R_{emp}(h_S^\phi, S^{\setminus i})]    
\end{align*}
We obtain the upper and lower bound for empirical risk using Equations \ref{eq:adj_gen_1} and \ref{eq:adj_gen_3} to get
\begin{align*}
\Ex_{\phi}[\ell(h_S^\phi, z_i)] - \Ex_{\phi}[\ell(h_{S^{\setminus i}}^{\phi}, z_j)] &\leq m\beta + 2m\gamma + (m-1)\Ex_{\phi}[R(h_{S^{\setminus i}}^{\phi}) + \gamma  - R(h_S^\phi) + \gamma]  \\
\Ex_{\phi}[\ell(h_S^\phi, z_i) - \ell(h_{S^{\setminus i}}^{\phi}, z_j)] &\leq m\beta + (4m-1)\gamma + (m-1)\Ex_{\phi}[R(h_{S^{\setminus i}}^{\phi})  - R(h_S^\phi) ]
\end{align*}
We add an subtract the risk of $h^* = \Ex[y| x]$ which is the true conditional of $\mathcal{D}^m$
\begin{align*}
\Ex_{\phi}[\ell(h_S^\phi, z_i) - \ell(h_{S^{\setminus i}}^{\phi}, z_j)] &\leq m\beta + (4m-1)\gamma + (m-1)\Ex_{\phi}[R(h_{S^{\setminus i}}^{\phi})  - R(h_S^\phi) \pm R(h^*)]\\
\Ex_{\phi}[\ell(h_S^\phi, z_i) - \ell(h_{S^{\setminus i}}^{\phi}, z_j)] &\leq m\beta + (4m-1)\gamma + (m-1)\Ex_{\phi}[R(h_{S^{\setminus i}}^{\phi}) - R(h^*) - (R(h_S^\phi) - R(h^*)) ]
\end{align*}
Using the uniform model bias bound from assumption \ref{as:model_bias} we have
\begin{align*}
\Ex_{\phi}[R(h_{S^{\setminus i}}^{\phi}) - R(h^*)] \leq \Delta\\
\Ex_{\phi}[R(h_S^\phi) - R(h^*)] \geq -\Delta
\end{align*}
Hence we get
\begin{align*}
\Ex_{\phi}[\ell(h_S^\phi, z_i) - \ell(h_{S^{\setminus i}}^{\phi}, z_j)] &\leq m\beta + (4m-1)\gamma + (m-1)\left[\Delta - (-\Delta) \right]\\
\Ex_{\phi}[\ell(h_S^\phi, z_i) - \ell(h_{S^{\setminus i}}^{\phi}, z_j)] &\leq m\beta + (4m-1)\gamma + 2(m-1)\Delta
\end{align*}
Since we can interchange $\ell(h_S^\phi, z_i)$ and $\ell(h_{S^{\setminus i}}^{\phi}, z_j)$ i.e. start with $\Ex_{z\sim \mathcal{D}}[\ell(h_{S^{\setminus i}}^{\phi}, z)] - \Ex_{z \sim \mathcal{D}}[\ell(h_S^\phi, z)]$ we have the result
\begin{align*}
\left\lvert \Ex_{\phi}[\ell(h_S^\phi, z_i) - \ell(h_{S^{\setminus i}}^{\phi}, z_j)] \right\rvert &\leq m\beta + (4m-1)\gamma + 2(m-1)\Delta \quad \blacksquare
\end{align*}

\begin{lemma} If Lipschitz assumption \ref{as:hess_lip} on the Hessian of $\ell$ holds from \citet{nesterov2006cubic} we have
\begin{align}
|\ell(h, z_1) - \ell(h, z_2) - \langle \nabla \ell(h, z_2), z_1 - z_2\rangle  - \langle \nabla^2 \ell(h, z_2)(z_1 - z_2), z_1 - z_2\rangle | \leq \cfrac{\rho}{6} |z_1 - z_2|^3 
\end{align}
\label{lm:hess_lip}
\end{lemma}

\subsection{Proof of Theorem \ref{th:mem_curv}}
\label{sec:proof_mem_curv}
From Lemma \ref{lm:hess_lip} we have
\begin{align*}
-\cfrac{\rho}{6} |z_1 - z_2|^3  \leq \ell(h, z_1) - \ell(h, z_2) - \langle \nabla \ell(h, z_2), z_1 - z_2\rangle  - \langle \nabla^2 \ell(h, z_2)(z_1 - z_2), z_1 - z_2\rangle  \leq \cfrac{\rho}{6} |z_1 - z_2|^3
\end{align*}

This gives us an upper bound on $\ell(h, z_1)$   
\begin{align}
\ell(h, z_1)  \leq \cfrac{\rho}{6} |z_1 - z_2|^3 + \ell(h, z_2) + \langle \nabla \ell(h, z_2), z_1 - z_2\rangle  + \langle \nabla^2 \ell(h, z_2)(z_1 - z_2), z_1 - z_2\rangle   \label{eq:loss_hess_upper}
\end{align}

Consider $z_j \in S$ such that $z_j = z_i + \alpha$ for some $j\neq i$ where $\alpha \in B_p(\upsilon)$ such that $\Ex[\alpha] = 0$ and $\Ex[\alpha^T \alpha] = 1$.

Without loss of generality from Lemma \ref{lm:loss_diff} we have:
\begin{align*}
\Ex_{\phi}[\ell(h_S^\phi, z_i) - \ell(h_{S^{\setminus i}}^{\phi}, z_j)] &\leq m\beta + (4m-1)\gamma + 2(m-1)\Delta
\end{align*}
Using the upper bound from Equation \ref{eq:loss_hess_upper}, setting $z_1 = z_j, z_2 = z_i$ we have
\begin{align*}
\Ex_{\phi}[\ell(h_S^\phi, z_i)  - \ell(h_{S^{\setminus i}}^{\phi}, z_j) - \cfrac{\rho}{6} \lVert \alpha \rVert^3 - \langle \nabla \ell(h_{S^{\setminus i}}^{\phi}, z_i), \alpha\rangle - \langle \nabla^2 \ell(h_{S^{\setminus i}}^{\phi}, z_i)\alpha, \alpha\rangle ] &\leq m\beta + (4m-1)\gamma + 2(m-1)\Delta \\
\Ex_{\phi}[\ell(h_S^\phi, z_i)  - \ell(h_{S^{\setminus i}}^{\phi}, z_i) - \cfrac{\rho}{6} \lVert \alpha \rVert^3 - \langle \nabla \ell(h_{S^{\setminus i}}^{\phi}, z_i), \alpha\rangle - \alpha^T H^T \alpha]  &\leq m\beta + (4m-1)\gamma + 2(m-1)\Delta
\end{align*}
Where $H =\nabla^2 \ell(h_{S^{\setminus i}}^{\phi}, z_i) $. Next, we take expectation over $\alpha$ we get
\begin{align*}
  \Ex_{\phi}[\ell(h_S^\phi, z_i)] & - \Ex_{\phi}[\ell(h_{S^{\setminus i}}^{\phi}, z_i)] - \cfrac{\rho}{6} \Ex_\alpha[\lVert \alpha \rVert^3] - \Ex_{\phi, \alpha}[\alpha^T H^T \alpha]  \leq m\beta + (4m-1)\gamma + 2(m-1)\Delta  \\
  \Ex_{\phi}[\ell(h_S^\phi, z_i)] & - \Ex_{\phi}[\ell(h_{S^{\setminus i}}^{\phi}, z_i)] - \cfrac{\rho}{6} \Ex_\alpha[\lVert \alpha \rVert^3] - \Ex_{\phi, \alpha}[\mathrm{tr}(H^T\Ex_\alpha[\alpha^T \alpha])]  \leq m\beta + (4m-1)\gamma + 2(m-1)\Delta  \\
  \Ex_{\phi}[\ell(h_S^\phi, z_i)] & - \Ex_{\phi}[\ell(h_{S^{\setminus i}}^{\phi}, z_i)] \leq \cfrac{\rho}{6} \Ex_\alpha[\lVert \alpha \rVert^3] + \Ex_{\phi}[\mathrm{tr}(H^T\Ex_\alpha[\alpha^T \alpha])]  + m\beta + (4m-1)\gamma + 2(m-1)\Delta  \\
  \Ex_{\phi}[\ell(h_S^\phi, z_i)] & - \Ex_{\phi}[\ell(h_{S^{\setminus i}}^{\phi}, z_i)] \leq \cfrac{\rho}{6} \Ex_\alpha[\lVert \alpha \rVert^3] + \Ex_{\phi}[\mathrm{tr}(H)] + m\beta + (4m-1)\gamma + 2(m-1)\Delta  \\
  \Ex_{\phi}[\ell(h_S^\phi, z_i)] & - \Ex_{\phi}[\ell(h_{S^{\setminus i}}^{\phi}, z_i)] \leq \cfrac{\rho}{6} \Ex_\alpha[\lVert \alpha \rVert^3] + \Ex_{\phi}[\mathrm{tr}(\nabla^2 \ell(h_{S^{\setminus i}}^{\phi}, z_i))] + m\beta + (4m-1)\gamma + 2(m-1)\Delta 
\end{align*}
If we have $0 \leq \ell \leq L$ then:
\begin{align}
   \cfrac{1}{L} \left[ \Ex_{\phi}[\ell(h_S^\phi, z_i)]  - \Ex_{\phi}[\ell(h_{S^{\setminus i}}^{\phi}, z_i)] \right] = \mathrm{mem}(\mathcal{A}, S, i)
\label{eq:mem_loss_diff_link}
\end{align}
Since we can exchange $S$ and $S^{\setminus i}$ Hence we have the result
\begin{align*}
    |\mathrm{mem}(\mathcal{A}, S, i)| \leq \cfrac{\rho}{6L}  \Ex_\alpha[\lVert \alpha \rVert^3] + \cfrac{1}{L}\Ex_{\phi}[\mathrm{tr}(\nabla^2 \ell(h_{S^{\setminus i}}^{\phi}, z_i))] + \cfrac{m\beta}{L} +  \cfrac{(4m-1)\gamma}{L} + \cfrac{2(m-1)\Delta}{L}  \quad \blacksquare
\end{align*}

\subsection{Proof of Theorem \ref{th:mem_curv} for Cross-Entropy}
\label{sec:proof_mem_curv_xe}

For classification with one-hot ground truth labels we have cross entropy we have. 
\begin{align*}
    \ell(h_S^\phi, z_i) &= - \ln(\Pr[h_S^\phi(x_i) = y_i]) \\
    \ell(h_S^\phi, z_i) - \ell(h_{S^{\setminus i}}^\phi, z_j) &= \ln \left( \cfrac{\Pr[h_{S^{\setminus i}}^\phi(x_j) = y_j]}{\Pr[h_S^\phi(x_i) = y_i]} \right)\\
    &= \ln \left(\cfrac{a}{b} \right)\\
    \text{for} \quad  \theta > -1 \quad& \text{we have,} \quad \frac{\theta}{\theta+1} \leq \ln(1+\theta)\\
    &\cfrac{\cfrac{a}{b}-1}{\cfrac{a}{b}} \leq \ln \left(\cfrac{a}{b} \right) \\
    &\cfrac{a-b}{a} \leq \ln \left(\cfrac{a}{b} \right) \\
    a - b \leq&\cfrac{a-b}{a} \leq \ln \left(\cfrac{a}{b} \right) \quad \text{For} ~ 0 < a \leq 1
\end{align*}
Thus we have
\begin{align*}
    \Pr[h_{S^{\setminus i}}^\phi(x_j) = y_j] -  \Pr[h_S^\phi(x_i) = y_i] \leq \ell(h_S^\phi, z_i) - \ell(h_{S^{\setminus i}}^\phi, z_j)
\end{align*}
Taking expectation over the randomness of $\mathcal{A}$ we have
\begin{align*}
    \Ex_{\phi}[ \Pr[h_{S^{\setminus i}}^\phi(x_j) = y_j]]-  \Ex_{\phi}[\Pr[h_S^\phi(x_i) = y_i]] &\leq  \Ex_{\phi}[\ell(h_S^\phi, z_i)] - \Ex_{\phi}[\ell(h^\phi_{S^{\setminus i}}, z_j)]\\
    \mathrm{mem}(\mathcal{A}, S, i) &\leq \Ex_{\phi}[\ell(h_S^\phi, z_i)] - \Ex_{\phi}[\ell(h^\phi_{S^{\setminus i}}, z_j)]  \quad \blacksquare
\end{align*}

\newpage
\subsection{Proof of Lemma \ref{lm:stab_priv}}
\label{sec:proof_stab_priv}

Let $h \sim \mathcal{A}(\phi, S)$ have a pdf defined as $p(h)$, and $h' \sim \mathcal{A}(\phi, S^{\setminus i})$ have a pdf defined as $p'(h')$
\begin{align*}
\left\lvert \Ex_{\phi, z}[\ell(h_S^\phi, z)] - \Ex_{\phi, z}[\ell(h^{\phi}_{S^{\setminus i}}, z)] \right\rvert
&=\left\lvert\Ex_{z,\phi}[\ell(h_S^\phi, z)] - \Ex_{z,\phi}[\ell(h^{\phi}_{S^{\setminus i}}, z)]\right\rvert\\
&=\left\lvert\Ex_{z,\phi}[\ell(\mathcal{A}(\phi, S), z)] - \Ex_{z,\phi}[\ell(\mathcal{A}(\phi, S^{\setminus i}), z)]\right\rvert\\
&=\left\lvert\Ex_{z, h}[\ell(h, z)] - \Ex_{z,h'}[\ell(h', z)]\right\rvert\\
&=\left\lvert\int_{z}\int_{h} \ell(h, z)p(h)dh~p(z)dz - \int_{z}\int_{h'} \ell(h', z)p'(h')dh'~p(z)dz\right\rvert\\
&=\left\lvert\int_{z}\int_{h} \ell(h, z)p(h)dh~p(z)dz - \int_{z}\int_{h} \ell(h, z)p'(h)dh~p(z)dz\right\rvert\\
&=\left\lvert\int_{z}\int_{h} \ell(h, z)(p(h)- p'(h))dh~p(z)dz\right\rvert\\
&\leq \left\lvert \int_{z}\sup_h \ell(h, z)\int_{h :p(h) \geq p'(h) }(p(h)- p'(h))dh~p(z)dz\right\rvert\\
&\leq \left\lvert \sup_{h,z} \ell(h, z)\int_{z}p(z)dz\int_{h  :p(h) \geq p'(h)}(p(h)- p'(h))dh\right\rvert\\
&\leq \left\lvert L\int_{h:p(h) \geq p'(h)}p(h)- p'(h)dh\right\rvert\\
&\leq \left\lvert L\int_{h:p(h) \geq p'(h)}p(h)\left(1- \cfrac{p'(h)}{p(h)}\right)dh\right\rvert\\
&\leq \left\lvert L \int_{h:p(h) \geq p'(h)}p(h)\left(1- e^{-\epsilon}\right)dh\right\rvert\\
&\leq \left\lvert L (1- e^{-\epsilon}) \int_{h:p(h) \geq p'(h)}p(h)dh\right\rvert\\
&\leq \left\lvert L (1- e^{-\epsilon})\right\rvert\\
&\leq  L (1- e^{-\epsilon}) \quad \blacksquare
\end{align*}

\begin{lemma}  If the assumptions of error stability \ref{as:stability}, generalization \ref{as:gen}, and uniform model bias \ref{as:model_bias} hold, then for two adjacent datasets $S, S^{\setminus i} \sim \mathcal{D}$ and for any $i, j \in \{1, \cdots, m \}$ with a probability at least $1-\delta$ we have
\label{lm:curv_intermidiate}
\begin{align}
    \Ex_{\phi}[\mathrm{tr}(\nabla^2 \ell(h^\phi_S, z_i))] \leq&~m\beta + (4m-1)\gamma \nonumber \\
    &+ 2(m-1)\Delta + \cfrac{\rho}{6} \Ex[\lVert \alpha \rVert^3] \nonumber\\
    &+ \Ex_\phi[\ell(h_S^\phi, z_j)]  - \Ex_\phi[\ell(h^\phi_{S^{\setminus i}}, z_j)].
\end{align}
\end{lemma}
\subsection{Proof of Lemma \ref{lm:curv_intermidiate}}
\label{sec:proof_curv_intermidiate}

From Lemma \ref{lm:hess_lip} we have
\begin{align*}
-\cfrac{\rho}{6} |z_1 - z_2|^3  \leq \ell(h, z_1) - \ell(h, z_2) - \langle \nabla \ell(h, z_2), z_1 - z_2\rangle  - \langle \nabla^2 \ell(h, z_2)(z_1 - z_2), z_1 - z_2\rangle  \leq \cfrac{\rho}{6} |z_1 - z_2|^3
\end{align*}
This gives us a lower bound on $\ell(h, z_1)$   
\begin{align}
-\cfrac{\rho}{6} |z_1 - z_2|^3 + \ell(h, z_2) + \langle \nabla \ell(h, z_2), z_1 - z_2\rangle  + \langle \nabla^2 \ell(h, z_2)(z_1 - z_2), z_1 - z_2\rangle &\leq \ell(h, z_1)   \label{eq:loss_hess_lower}
\end{align}
Consider $z_j \in S$ such that $z_i = z_j + \alpha$ for some $j\neq i$ where $\alpha \in B_p(\upsilon)$ such that $\Ex[\alpha] = 0$ and $\Ex[\alpha^T \alpha] = 1$. Using the lower bound in Lemma \ref{lm:loss_diff} with $z_1 = z_i, z_2 = z_j$ we get

\begin{align*}
\Ex_\phi[\ell(h^\phi_S, z_i)] - \Ex_\phi[\ell(h_{S^{\setminus i}}^\phi, z_j)] &\leq m\beta + (4m-1)\gamma + 2(m-1)\Delta  \\
- \cfrac{\rho}{6} \lVert\alpha\rVert^3 + \Ex_\phi[\ell(h_S^\phi, z_j)] + \Ex_\phi[\langle \nabla \ell(h_S^\phi, z_j), \alpha\rangle] + \Ex_\phi[\langle \nabla^2 \ell(h_S^\phi, z_i)\alpha, \alpha\rangle]  - \Ex_\phi[\ell(h^\phi_{S^{\setminus i}}, z_j)] &\leq m\beta + (4m-1)\gamma + 2(m-1)\Delta \\
\Ex_\phi[\ell(h_S^\phi, z_j)] + \Ex_\phi[\langle \nabla \ell(h_S^\phi, z_j), \alpha\rangle] + \Ex_\phi[\langle \nabla^2 \ell(h_S^\phi, z_i)\alpha, \alpha\rangle]  - \Ex_\phi[\ell(h^\phi_{S^{\setminus i}}, z_j)]  \leq m\beta &+ (4m-1)\gamma + 2(m-1)\Delta + \cfrac{\rho}{6} \lVert \alpha \rVert^3 
\end{align*}
Taking Expectation over $\alpha$ we get
\begin{align*}
\Ex_\phi[\ell(h_S^\phi, z_j)] + \Ex_{\alpha, \phi}[\langle \nabla \ell(h_S^\phi, z_j), \alpha\rangle] + \Ex_{\alpha,\phi}[\langle \nabla^2 \ell(h_S^\phi, z_i)\alpha, \alpha\rangle]  - \Ex_\phi[\ell(h^\phi_{S^{\setminus i}}, z_j)]  &\leq m\beta + (4m-1)\gamma + 2(m-1)\Delta + \cfrac{\rho}{6} \lVert \alpha \rVert^3
\end{align*}
Note that we can change the order of expectation due to Fubini's theorem
\begin{align*}
\Ex_\phi[\ell(h_S^\phi, z_j)] + \Ex_{\phi, \alpha}[\langle \nabla^2 \ell(h_S^\phi, z_i)\alpha, \alpha\rangle]  - \Ex_\phi[\ell(h^\phi_{S^{\setminus i}}, z_j)]  &\leq m\beta + (4m-1)\gamma + 2(m-1)\Delta + \cfrac{\rho}{6} \lVert \alpha \rVert^3 \\
\Ex_\phi[\ell(h_S^\phi, z_j)] + \Ex_{\phi}[\mathrm{tr}(\nabla^2 \ell(h^\phi_S, z_i))]  - \Ex_\phi[\ell(h^\phi_{S^{\setminus i}}, z_j)] &\leq m\beta + (4m-1)\gamma + 2(m-1)\Delta + \cfrac{\rho}{6} \lVert \alpha \rVert^3 \\
\Ex_{\phi}[\mathrm{tr}(\nabla^2 \ell(h^\phi_S, z_j))] \leq m\beta + (4m-1)\gamma + 2(m-1)\Delta +& \cfrac{\rho}{6} \Ex[\lVert \alpha \rVert^3] + \Ex_\phi[\ell(h_S^\phi, z_j)]  - \Ex_\phi[\ell(h^\phi_{S^{\setminus i}}, z_j)] \quad \blacksquare
\end{align*}

\subsection{Proof of Theorem \ref{th:pr_curv}}
\label{sec:proof_pr_curv}
We start with the results of Lemma \ref{lm:curv_intermidiate}. Taking Expectation over $z \sim \mathcal{D}$ we have
\begin{align*}
\Ex_{z, \phi}[\mathrm{tr}(\nabla^2 \ell(h^\phi_S, z))]  &\leq m\beta + (4m-1)\gamma + 2(m-1)\Delta + \cfrac{\rho}{6} \Ex[\lVert \alpha \rVert^3] + \Ex_{z, \phi}[\ell(h_S^\phi, z_j)]  - \Ex_{z, \phi}[\ell(h^\phi_{S^{\setminus i}}, z_j)]  \\
\Ex_{z,\phi}[\mathrm{tr}(\nabla^2 \ell(h^\phi_S, z))]    &\leq m\beta + (4m-1)\gamma + 2(m-1)\Delta + \cfrac{\rho}{6} \Ex[\lVert \alpha \rVert^3] + \beta \\
\Ex_{z, \phi}[\mathrm{tr}(\nabla^2 \ell(h^\phi_S, z))]    &\leq (m+1)\beta + (4m-1)\gamma + 2(m-1)\Delta + \cfrac{\rho}{6} \Ex[\lVert \alpha \rVert^3]
\end{align*}
Using Lemma \ref{lm:stab_priv}
\begin{align*}
\Ex_{z, \phi}[\mathrm{tr}(\nabla^2 \ell(h^\phi_S, z))]    &\leq L(m+1)(1 - e^{-\epsilon}) + (4m-1)\gamma + 2(m-1)\Delta + \cfrac{\rho}{6} \Ex[\lVert \alpha \rVert^3] \quad \blacksquare
\end{align*}

\end{document}